\documentclass[superscriptaddress]{revtex4}

\usepackage{color}
\usepackage{graphics}
\usepackage[pdftex]{graphicx}
\usepackage{amsmath}
\usepackage{amsfonts}
\usepackage{float}
\usepackage{natbib}
\usepackage{epstopdf}
\usepackage{verbatim}
\usepackage{algorithm}
\usepackage{algorithmic}
\usepackage{todonotes}
\usepackage{lineno}
\usepackage{bbold}
\usepackage{array}
\usepackage{mathrsfs}
\usepackage{amsbsy}
\usepackage{mathtools}
\usepackage{chngcntr}
\usepackage{latexsym}
\usepackage{microtype}

\usepackage{hyperref}
\hypersetup{
    colorlinks=true,
    citecolor={blue!100!black},
    linkcolor={red!100!black},
    urlcolor={blue!80!black}
}

\usepackage{pdfpages}

\begin{document}

\title{\Large{Elastica: A compliant mechanics environment for soft robotic control}}

\author{Noel Naughton}
\affiliation{Department of Mechanical Science and Engineering, University of Illinois at Urbana-Champaign}
\affiliation{Coordinated Science Laboratory, University of Illinois at Urbana-Champaign}

\author {Jiarui Sun}
\affiliation{Department of Electrical and Computer Engineering, University of Illinois at Urbana-Champaign}

\author{Arman Tekinalp}
\affiliation{Department of Mechanical Science and Engineering, University of Illinois at Urbana-Champaign}

\author{Girish Chowdhary}
\affiliation{Coordinated Science Laboratory, University of Illinois at Urbana-Champaign}
\affiliation{Department of Agricultural and Biological Engineering, University of Illinois at Urbana-Champaign}
\affiliation{Department of Computer Science, University of Illinois at Urbana-Champaign}

\author{Mattia Gazzola} 
\email{mgazzola@illinois.edu}
\affiliation{Department of Mechanical Science and Engineering, University of Illinois at Urbana-Champaign}
\affiliation{National Center for Supercomputing Applications, University of Illinois at Urbana-Champaign}
\affiliation{Carl R. Woese Institute for Genomic Biology, University of Illinois at Urbana-Champaign}
\affiliation{Center for Artificial Intelligence Innovation, University of Illinois at Urbana-Champaign}

\begin{abstract}
\vspace{5pt}
\begin{center}
\textbf{ABSTRACT}
\vspace{-5pt}
\end{center}
Soft robots are notoriously hard to control. This is partly due to the scarcity of models able to capture their complex continuum mechanics, resulting in a lack of control methodologies that take full advantage of body compliance. Currently available simulation methods are either too computational demanding or overly simplistic in their physical assumptions, leading to a paucity of available simulation resources for developing such control schemes. To address this, we introduce Elastica, a free, open-source simulation environment for soft, slender rods that can bend, twist, shear and stretch. We demonstrate how Elastica can be coupled with five state-of-the-art reinforcement learning algorithms to successfully control a soft, compliant robotic arm and complete increasingly challenging tasks. 
\end{abstract}


\maketitle

\vspace{-10pt}




\section{Introduction}

The introduction of soft materials in robotics has long been seen as key to access capabilities that are new or complementary to traditionally rigid structures via enhanced dexterity, safety, versatility, and adaptability, with opportunities in industry, agriculture, assistive care, medicine, and defense \cite{cianchetti2018biomedical, polygerinos2017soft,chowdhary2019soft}.


A major challenge in fulfilling this potential is that soft robots are notoriously hard to control \cite{rus2015design,george2018control}. While part of the problem is related to material and fabrication constraints, from an algorithmic perspective two main aspects distinctly set them apart from rigid-body robots. First, the controller needs to orchestrate virtually infinite degrees of freedom via a finite set of actuators. This renders soft robots characteristically hyper-redundant and underactuated \cite{trivedi2008soft}. Second, these continuum systems are subject to highly non-linear and long range stress propagation effects. As a consequence, localized loads are communicated throughout the entire structure, potentially inducing global (and sometimes dramatic) shape reconfigurations \cite{Charles:2019}. Thus, control strategies for compliant robots are inextricably connected to their complex physics: failure to model and capture it often amounts to failing at control. 

The current lack of rigorous, accurate and efficient numerical models able to account for the mechanics at play contributes to further strain the control issue. On the other hand, if compliant effects, modes, and instabilities can be faithfully captured, then there is the opportunity to take advantage of them to, in fact, \textit{simplify} the control problem. Indeed, it has been shown that this `mechanical intelligence' \cite{Pfeifer:2007} synthesized by elastic modes can be leveraged to coordinate complex locomotion behaviors \cite{Gazzola:2015} as well as topological transitions that can be harnessed for work \cite{Charles:2019}.


Along these lines, compliance allows us to think of obstacles and boundaries as potential allies. In the case of robotic arms and manipulators, solid interfaces are classically dealt with through additional constraints or penalties that render the control problem harder to solve \cite{khatib1986real,chang2020energy}. This active obstacle avoidance strategy is justified in rigid-link robots by the fact that impacts with obstacles can cause damage and to prevent geometric frustration and locking into undesired poses. In contrast, compliant robots can safely conform to, and therefore \textit{exploit}, solid boundaries to correct imprecise actuation, re-distribute excessive loads, or favorably reshape themselves. This contrast may be intuitively summarized as avoiding obstacles versus \textit{leaning} against them. 

In order to facilitate the exploration of these concepts and development of control strategies for soft, slender robotic structures, we introduce here Elastica, a free, open-source simulation environment specifically tailored to the soft robotic context. Elastica's physics engine implements a methodology based on Cosserat rods \cite{Gazzola2018}, which are slender, three-dimensional, dynamic, and continuum elements that can bend, twist, shear, and stretch at every cross-section. Their clean mathematical formulation naturally accommodates environmental loads, making them particularly attractive for modeling interface effects such as contact, self-contact, friction, or hydrodynamics. Hence, they are well suited to aid the development of robotic arm counterparts that are soft, flexible, and tailored to reaching and manipulation tasks in unstructured, dynamic environments \cite{trivedi2008soft,walker2005continuum,calisti2011octopus}. 

Our approach aims at filling the gap between conventional spring-and-damper rigid body solvers that are unable to capture the second-order effects responsible for characteristic elastic modes, and high-fidelity finite elements methods (FEM), which are mathematically cumbersome and often prohibitively expensive. Elastica's methods have been shown to strike a valuable compromise between these two approaches, and their practical utility has been demonstrated in a number of engineering and biophysical contexts that encompass both individual and complex assemblies of Cosserat rods: from design and fabrication of bio-hybrid soft robots made of muscle tissue \cite{Wang:bioRxiv,pagan2018simulation}, neurons, and artificial scaffolds \cite{aydin2019neuromuscular}, to the computational replica of intricate biological systems that include flexing human elbow joints, slithering snakes, and flapping feathered wings \cite{Zhang2019}. 

Here, Elastica is implemented to interface with major reinforcement learning (RL) packages such as Stable Baselines \cite{stable-baselines}, illustrating the coupling of physics and control. We demonstrate the ability of popular state-of-the-art RL methods (TRPO, PPO, SAC, DDPG, and TD3) to deal with increasingly challenging scenarios in which an actor learns to control a soft arm's modes of deformation to track and reach a target, as well as maneuver around obstacles. Our goal is not necessarily to establish RL as the method of choice for this class of problems, but to illustrate how Elastica enables benchmarking and development of control methods in a soft mechanics context.

Overall, our results confirm the successful coupling of RL with Elastica to effectively carry out challenging control tasks and practically illustrate how compliant mechanics and solid boundaries can be used to our advantage. The software interfaces provided by Elastica allow the user to tap into well-developed control libraries as well as easily define control tasks, variables, actuation modalities, and physical environments, establishing Elastica as a useful testing ground for control methods, complementing popular platforms such as PyBullet \cite{PyBullet} or MuJoCo \cite{MuJoCo}.


\section{Related Work}

\textbf{Physical simulation environments.} Because of the unique physics of soft, compliant robots, RL agents must be trained using special-purpose simulation frameworks \cite{Bhagat_2019}. Current simulation environments typically used for RL, such as PyBullet, \cite{PyBullet}, DART \cite{DART}, and MuJoCo \cite{MuJoCo}, simulate multi-joint dynamics via efficient recursive algorithms combined with modern velocity-stepping methods for contact dynamics. These methods capture the dynamics of rigid robots and similar approaches have been applied to control multi-link arms inspired by octopuses \cite{engel2006learning}. However, these simulation methods intrinsically fail to capture higher-order continuum elastic effects and associated dynamics, limiting the ability of a given control scheme to fully exploit available deformation modes.

\textbf{Modeling of continuum robots.} In a robotic context characterized by large deformations in 3D space, non-linear mechanics, continuous actuation, and interface effects, minimal theoretical models or first order approximations based on springs, dampers, and linkages \cite{Yamaguchi:2005} are ill-suited to fully capture the rich dynamics of intrinsically soft bodies. On the other side of the spectrum, high fidelity 3D FEMs have been used to simulate and design various soft robotic components \cite{coevoet2017software}. However, these methods also exhibit limitations such as often prohibitive computational costs, numerical instabilities, loss of accuracy due to mesh distortion, and involved mathematical representation. Consequently, the use of FEM has been relatively limited in the modeling of soft robots. 

Alternative approaches often seek to leverage geometric slenderness. Slender objects are then treated as one-dimensional elastic curves, significantly reducing mathematical complexity and computational costs, while retaining physical accuracy. The graphics community has been most active in this area, where spline-based strands \cite{Pai:2004} and discrete rod models \cite{Bergou:2008} (based on the unstretchable and unsharable Kirchhoff model \cite{Kirchhoff:1859}) are routinely used in a variety of realistic simulations, from elastic ribbons and woven cloth to entangled hair and human tendons. In robotics, similar methods have been used \cite{trivedi2008soft} to model soft arms \cite{Sadati:2019,Armanini:2017,Connolly:2017, uppalapati2020BerryPickingRobot}, snake robots \cite{Cicconofri:2015} or surgical manipulators \cite{Mahvash:2011}. Although numerically efficient, these approaches are specialized to scenarios in which either shear, stretch, twist, dynamic effects, or environmental loads are unimportant. Lately, there has been a need to generalize these models to explain phenomena related to artificial muscle coiling \cite{Gazzola2018,Charles:2019}, highly stretchable and shearable elastomers, or biological components integrated in robotic substrates \cite{Zhang2019}. Thus, the more comprehensive Cosserat rod models \cite{Cosserat:1909} have been gaining attention and recently demonstrated their utility in a range of applications, from bio-hybrid soft robotics to biophysics \cite{Zhang2019}.

\textbf{Reinforcement learning for soft robotic control.} 
Soft robots are difficult to control with traditional control methods due to their virtually infinite degrees of freedom and highly nonlinear continuum dynamics \cite{soft_survey, chin2020machine, chowdhary2019soft}. This has created fertile ground for using model-free reinforcement learning to control soft robots. For example, Satheeshbabu et al. \cite{satheeshbabu2019open, satheeshbabu2020continuous} present model-free approaches for position control of a soft spatial-continuum arm using variants of deep Q-learning (DQN) \cite{mnih2015humanlevel} and Deep Deterministic Policy Gradient (DDPG) \cite{lillicrap2015continuous}. Uppalapati et al. used DDPG based control of a hybrid rigid-soft arm and manipulator in cluttered agricultural environments \cite{uppalapati2020BerryPickingRobot}. Since these RL methods can be sample expensive, the authors in these papers used static, force-based simulations of the arm to train the RL policies and evaluated them on real soft robot arms. The resulting policies demonstrate feasibility of the RL approach for soft robotics, but the limiting assumptions of the models restrict the real world performance of the learned policies. We expect the realistic, continuum mechanics based models implemented in this paper will enable soft robotics researchers to learn better control policies in simulation. We note that our goal is not necessarily to establish model-free RL as the method of choice for these systems. Nonetheless, our results do show it to be a useful and convenient option. Indeed, complex mecahnics, 3D effects, and obstacles make it difficult to derive suitable (semi-)analytical descriptions that can be employed in model-based techniques. 


\begin{figure}[t!]
\centering
\includegraphics[width=\textwidth]{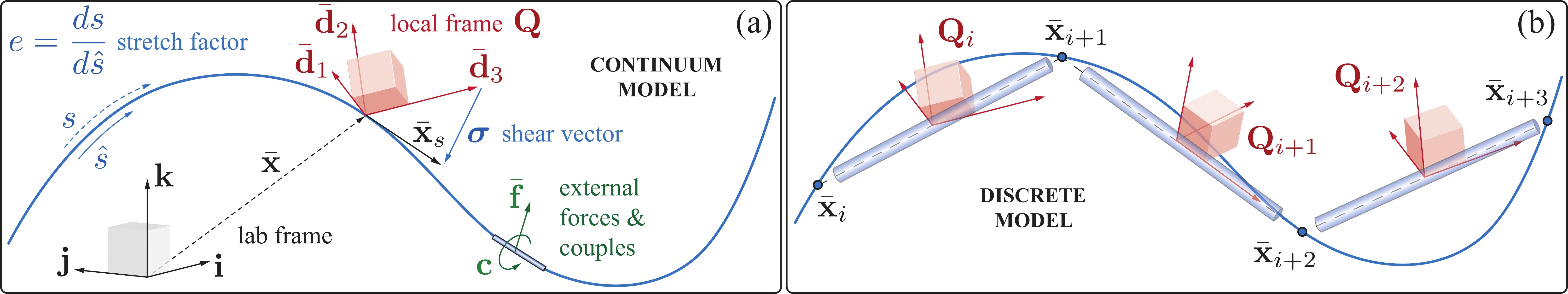}
\caption{Cosserat rods. (a) Continuum model and (b) discretization in Elastica \cite{Gazzola2018,Zhang2019}.}
\label{fig:elastica}
\vspace{-10pt}
\end{figure}

\section{Computational environment}
\label{sec:methods}


\textbf{Cosserat rod model.} Based on Cosserat rod theory \cite{Cosserat:1909}, we describe a rod (slender body, Fig.~\ref{fig:elastica}a) by a centerline $\bar{\mathbf{x}}(s, t) \in \mathbb{R}^3$ and rotation matrix $\mathbf{Q}(s, t)=\{ \bar{\mathbf{d}}_1, \bar{\mathbf{d}}_2, \bar{\mathbf{d}}_3 \}^{-1}$ which leads to a general relation between frames for any vector $\mathbf{v}$: $\mathbf{v}=\mathbf{Q}\bar{\mathbf{v}}$, $\bar{\mathbf{v}}=\mathbf{Q}^T \mathbf{v}$, where $\bar{\mathbf{v}}$ denotes a vector in the lab frame and $\mathbf{v}$ denotes a vector in the local frame.  Here $s \in [0, L_0]$ is the material coordinate of a rod of rest-length $L_0$, $L$ denotes the deformed filament length and $t$ is time. If the rod is unsheared, $\bar{\mathbf{d}}_3$ points along the centerline tangent $\partial_s \bar{\mathbf{x}}=\bar{\mathbf{x}}_s$ while $\bar{\mathbf{d}}_1$ and $\bar{\mathbf{d}}_2$ span the normal-binormal plane. Shearing and extension shift $\bar{\mathbf{d}}_3$ away from $\bar{\mathbf{x}}_s$, which can be quantified with the shear vector $\boldsymbol{\sigma}=\mathbf{Q} (\bar{\mathbf{x}}_s-\bar{\mathbf{d}}_3) = \mathbf{Q}\bar{\mathbf{x}}_s - \mathbf{d}_3$ in the \emph{local} frame. The curvature vector $\boldsymbol{\kappa}$ encodes $\mathbf{Q}$'s rotation rate along the material coordinate $\partial_s \mathbf{d}_j = \boldsymbol{\kappa} \times \mathbf{d}_j$, while the angular velocity $\boldsymbol{\omega}$ is defined by $\partial_t \mathbf{d}_j = \boldsymbol{\omega} \times \mathbf{d}_j$. We also define the velocity of the centerline $\bar{\mathbf{v}} = \partial_t\bar{\mathbf{x}}$ and, in the rest configuration, the bending $\mathbf{B}$ and shearing $\mathbf{S}$ stiffness matrices, second area moment of inertia $\mathbf{I}$, cross-sectional area $A$ and mass per unit length $\rho$. Then, the dynamics of a slender body reads as
{\footnotesize
\begin{eqnarray}
\rho A \cdot \partial_t^2 \bar{\mathbf{x}} &=& \partial_s \left( \frac{\mathbf{Q}^T \mathbf{S} \boldsymbol{\sigma}}{e} \right) + e\bar{\mathbf{f}}\label{eq:lin}\\
\frac{\rho \mathbf{I}}{e} \cdot \partial_t \boldsymbol{\omega} &=& \partial_s \left( \frac{\mathbf{B} \boldsymbol{\kappa}}{e^3} \right) + \frac{\boldsymbol{\kappa} \times \mathbf{B} \boldsymbol{\kappa}}{e^3} + \left( \mathbf{Q}\frac{\bar{\mathbf{x}}_s}{e} \times \mathbf{S} \boldsymbol{\sigma} \right) + \left( \rho \mathbf{I} \cdot \frac{\boldsymbol{\omega}}{e} \right) \times \boldsymbol{\omega} + \frac{\rho \mathbf{I} \boldsymbol{\omega}}{e^2} \cdot \partial_t e + e\mathbf{c}\label{eq:ang}
\end{eqnarray}
\normalsize}
\noindent where Eqs.~(\ref{eq:lin}, \ref{eq:ang}) represent linear and angular momentum balance at every cross section, $e = |\bar{\mathbf{x}}_s|$ is the local stretching factor, and $\bar{\mathbf{f}}$ and $\mathbf{c}$ are the external force and couple line densities, respectively.

This representation entails a number of favorable features: \textit{(1)} it captures 3D dynamics accounting for all modes of deformation -- bend, twist, shear, and stretch; \textit{(2)} continuum actuation, interface effects and environmental loads can be directly combined with body dynamics via $\bar{\mathbf{f}}$ and $\mathbf{c}$, making their inclusion straightforward; \textit{(3)} its complexity scales linearly with axial resolution, compared to cubic for FEM, significantly reducing compute time. Discretization of the above system of equations, along with appropriate boundary conditions, allows modeling dynamics of multiple active or passive Cosserat rods interacting with each other and the environment. More details are available in \cite{Gazzola2018,Zhang2019}.

\textbf{Simulation and problems setup -- a compliant robotic arm.} A particularly promising area of soft robotics is the development of continuum, compliant arms capable of reaching and manipulation tasks in complex, dynamic environments. Often inspired by cephalopod arms and other muscular hydrostats such as elephant trunks \cite{trivedi2008soft,walker2005continuum,calisti2011octopus}, these hyper-redundant, compliant robots promise a host of advantages such as increased maneuverability, dexterity, and safety. These robots are particularly amenable to being represented within Elastica as they can be accurately modeled as single, slender rods. Here, we consider four different scenarios of increasing complexity in which an RL actor is tasked with learning to control a compliant arm simulated in Elastica.

In all cases the goal is for the tip of the arm to reach a target location, complemented by additional, case-specific requirements. \textit{Case 1:} tracking a randomly moving target in 3D space. \textit{Case 2:} reaching to a randomly located stationary point and orienting the arm so the tip of the arm matches a randomly prescribed target orientation. \textit{Case 3:} learning to interact with and exploit solid boundaries to enable underactuated maneuvering through structured obstacles. \textit{Case 4:} underactuated maneuvering through an unstructured nest of obstacles. In all cases, the reward function is defined so that an episode score above zero is indicative of (at least partially) successful completion of the task, with higher scores corresponding to faster and more consistent task completion.

The arm is modeled as a single Cosserat rod that is fixed upright at its base and free to move in 3D space. The arm has an elastic Young's modulus of 10 MPa, leading to a bending stiffness typical of soft robotic arms \cite{rus2015design}. Actuation of the arm is achieved via application of internal torques distributed along the length of the arm. These continuum activation functions are modeled via splines characterized by $N$ independent control points and vanishing values (i.e. zero couple) at the extrema of the arm \cite{Gazzola2018}. The arm is controlled by decomposing the overall actuation into orthogonal torque functions applied in the local normal and binormal directions (i.e. along ${\mathbf{d}}_1$ and ${\mathbf{d}}_2$), which cause the arm to bend, as well as in the orthonormal direction ${\mathbf{d}}_3$, which causes the arm to twist. Different actuation modes (only bending or bending/twisting) are provided in the various cases, as detailed in the relative sections. Specific details of the spline actuation method, the physical and numerical parameters, and descriptions of the action spaces, states, and reward for each case is in the SI .

\textbf{Selected RL methods.} To investigate the ability of RL to dynamically control a compliant robotic arm in Elastica, five model-free, policy-gradient RL methods were considered, consisting of two methods -- Trust Region Policy Optimization (TRPO) and Proximal Policy Optimization (PPO) -- implemented as on-policy and three off-policy algorithms -- Soft Actor Critic (SAC), Deep Deterministic Policy Gradient (DDPG) and Twin Delayed DDPG (TD3). These algorithms are considered to be some of the best currently available RL algorithms for continuous control, with demonstrated performance in a variety of tasks. We used implementations provided by the OpenAI-derived Stable Baselines library \cite{stable-baselines}. Limited hyperparameter tuning was performed with most parameters kept at default value. Details of the parameters used are available in the SI. While the lack of extensive hyperparameter tuning suggests the possibility that the scores reported here are not the maximum attainable for each algorithm, the purpose of this work is not to adjudicate which of the selected algorithms is the best at these particular cases, but rather demonstrate their utility in combination with Elastica, and to establish a baseline against which these, and other algorithms can be measured.


\begin{figure}[t!]
\centering
\includegraphics[width=\textwidth]{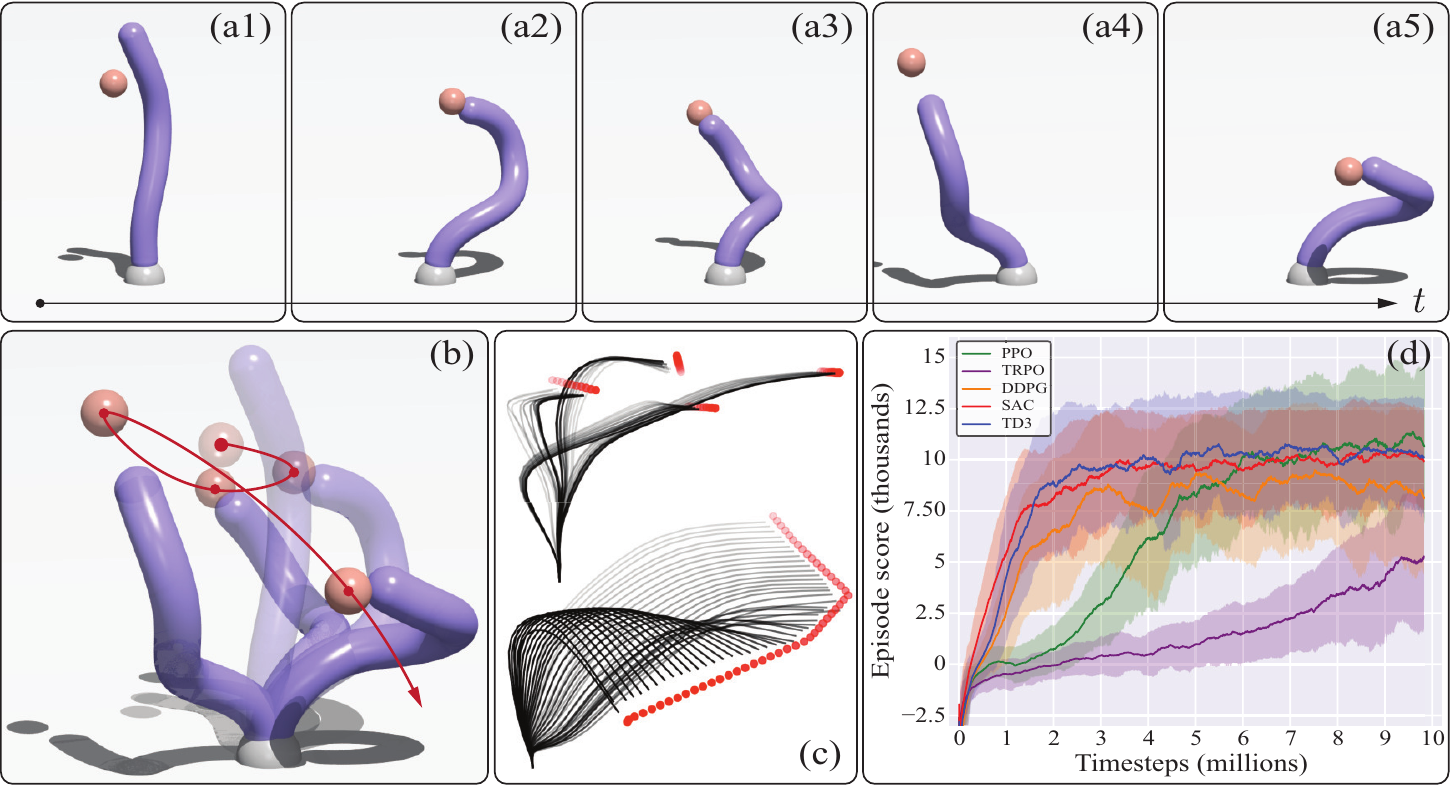}
\caption{a) Snapshots from a trained policy (SAC) over the course of one episode showing the arm successfully tracking a randomly moving target. b) Overlay of snapshots showing the random trajectory of the target. c) Trajectory of arm centerline and the target over successive timesteps. d) Learning results of the different algorithms. Algorithms were trained with 5 different random seeds for 10 millions timesteps. Curves are the rolling 250 sample average of the combined results and shaded regions are the standard deviation of the sample. }
\label{fig:prob1}
\vspace{-10pt}
\end{figure}

\section{Results}
\textbf{Case 1 -- 3D tracking of a randomly moving target.} The first case consists of the tip of the arm continuously tracking a randomly moving target in 3D space as illustrated in Figure \ref{fig:prob1}a-c. The reward function to be maximized consists of a penalty term proportional to the distance between the tip of the arm and the target combined with a two-tier bonus reward as the tip approaches the target. Actuation is allowed only in the normal and binormal directions (3D bending, but no twist). The actuation function in each direction is controlled by 6 equidistantly spaced control points leading to an action space with 12 degrees of freedom (DOF). The state contains the location of 11 points spaced equidistantly along the arm, the arm tip's velocity direction and magnitude, the target location, and the target's velocity direction and magnitude. Hyperparameter tuning was limited to the batch size (1000 to 128k) for on-policy methods (TRPO, PPO), and to the replay buffer size (100k to 2M) for off-policy methods (SAC, DDPG, TD3). Policies were trained for either 10 million (TRPO, PPO) or 5 million timesteps (SAC, DDPG, TD3). Hyperparameter tuning results are available in the SI.

For the on-policy methods, the best performance was achieved with a batch size of 8000 samples for TRPO and 32,000 for PPO. For the off-policy methods, all three methods achieved their best performance with a replay buffer size of 2 million samples. The learning curves for the best performing setups (and 10 millions timesteps training) are shown in Figure \ref{fig:prob1}d. All methods learn to track the moving target satisfactorily, although with differences. SAC, TD3, and PPO achieve similar scores, with SAC and TD3 converging faster than PPO. DDPG learns at a similar rate to SAC and TD3, but converges to a score $\sim$20\% lower. TRPO achieves the lowest score, though it appears to not have converged, so that with additional training comparable scores may be achieved. 

\begin{figure}[t!]
\centering
\includegraphics[width=\textwidth]{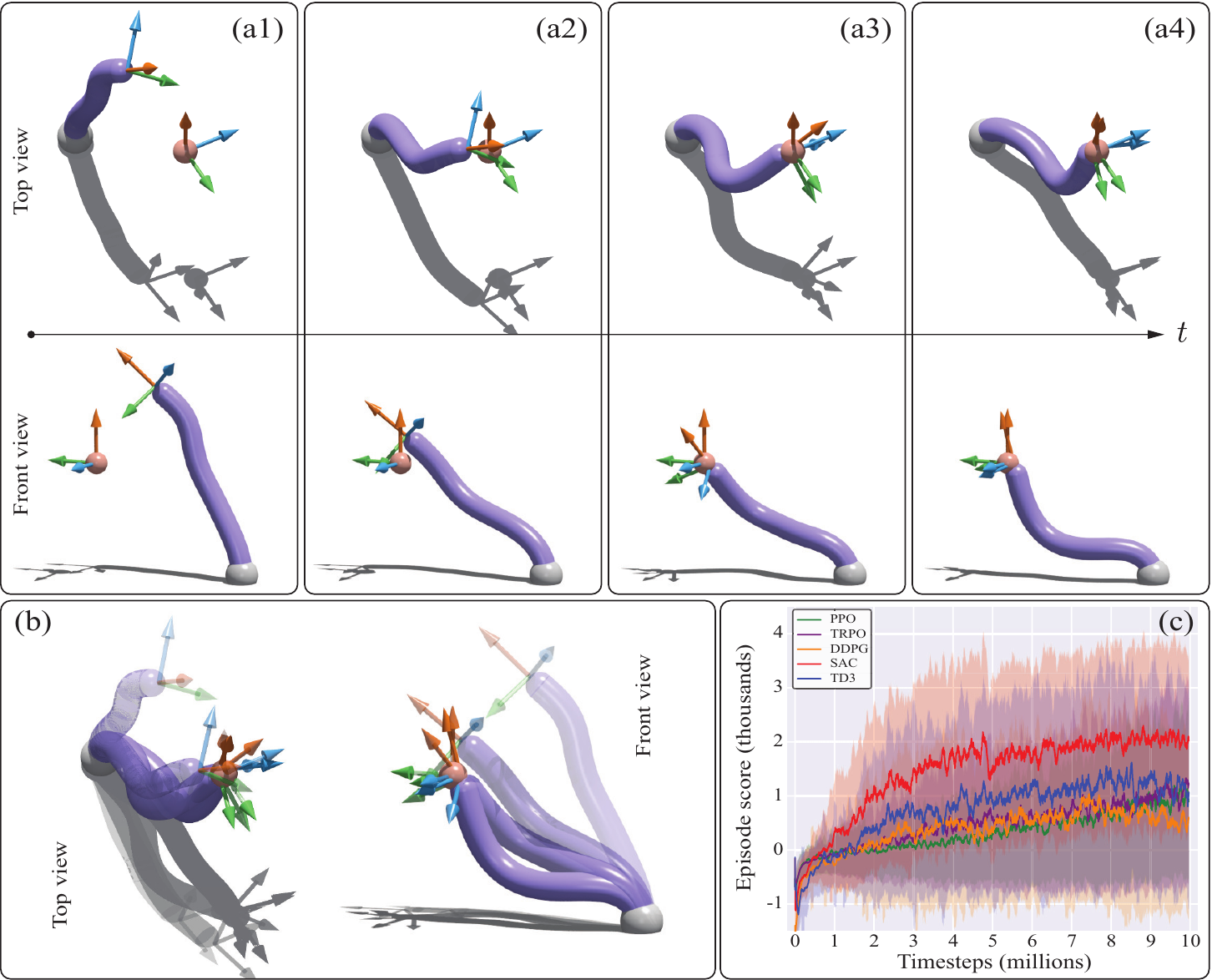}
\caption{a) Snapshots from a trained policy (SAC) over the course of one episode showing the arm successfully reaching the target and then orienting itself to match the target orientation. b) Overlay of the snapshots. c) Learning results of the different algorithms. Algorithms were trained with 5 different random seeds for 10 millions timesteps. Curves are the rolling 250 sample average of the combined results and shaded regions are the standard deviation of the sample.}
\label{fig:prob2}
\vspace{-10pt}
\end{figure}

\textbf{Case 2 -- Reaching to randomly located target with defined orientation.} Manipulation of objects by changing their orientation is a key use case of robotic arms. Case 2 consists of reaching to a randomly located, stationary target while re-shaping to match a desired end-effector orientation (Fig.~\ref{fig:prob2}a-b). The target coordinate frame is defined with the axial direction ($\bar{\mathbf{d}}_3$) pointing vertically upward and the normal-binormal directions ($\bar{\mathbf{d}}_1$, $\bar{\mathbf{d}}_2$) randomly rotated in-plane. The reward function is similar to Case 1 but with an additional penalty proportional to the difference between tip and target orientations and bonus rewards as the orientations become aligned. For the arm to match the prescribed orientation it is necessary to include twist as a mode of deformation. Similar to the bending actuation of Case 1, twist is controlled by 6 equidistantly distributed control points. Combining bending and twisting actuations yields an action space with 18 DOFs. The state information is that of Case 1 but with the addition of two quaternions to represents the arm tip's and target's orientation. 

Hyperparameter tuning was performed in the same manner as Case 1. The best performance is seen for a batchsize of 16,000 for both TRPO and PPO and, as in Case 1, the best replay buffer size for SAC, DDPG and TD3 is 2 million samples. The learning curves for the best performing algorithms are shown in Figure \ref{fig:prob2}c. All algorithms learn to at least partially complete the task, though SAC outperforms all algorithms with a final average score almost twice that of the others. All other algorithms exhibit similar performance. Notably, TRPO and PPO have similar performance, in contrast to PPO outperforming TRPO in Case 1. Finally, all algorithms exhibit a large variance, explained by the fact that not all target location/orientation pairs are physically attainable by the arm as well as a dependence on the selected random seed.

Case 1 and Case 2 demonstrate that the selected RL methods are capable of learning to control soft bodies in 3D space, and effectively manipulate their pose via distributed deformations modes generally not available to rigid counterparts, particularly twist. Next, we challenge these methods to learn how the arm can advantageously interact with its environment through the addition of obstacles. 

\begin{figure}[t!]
\centering
\includegraphics[width=\textwidth]{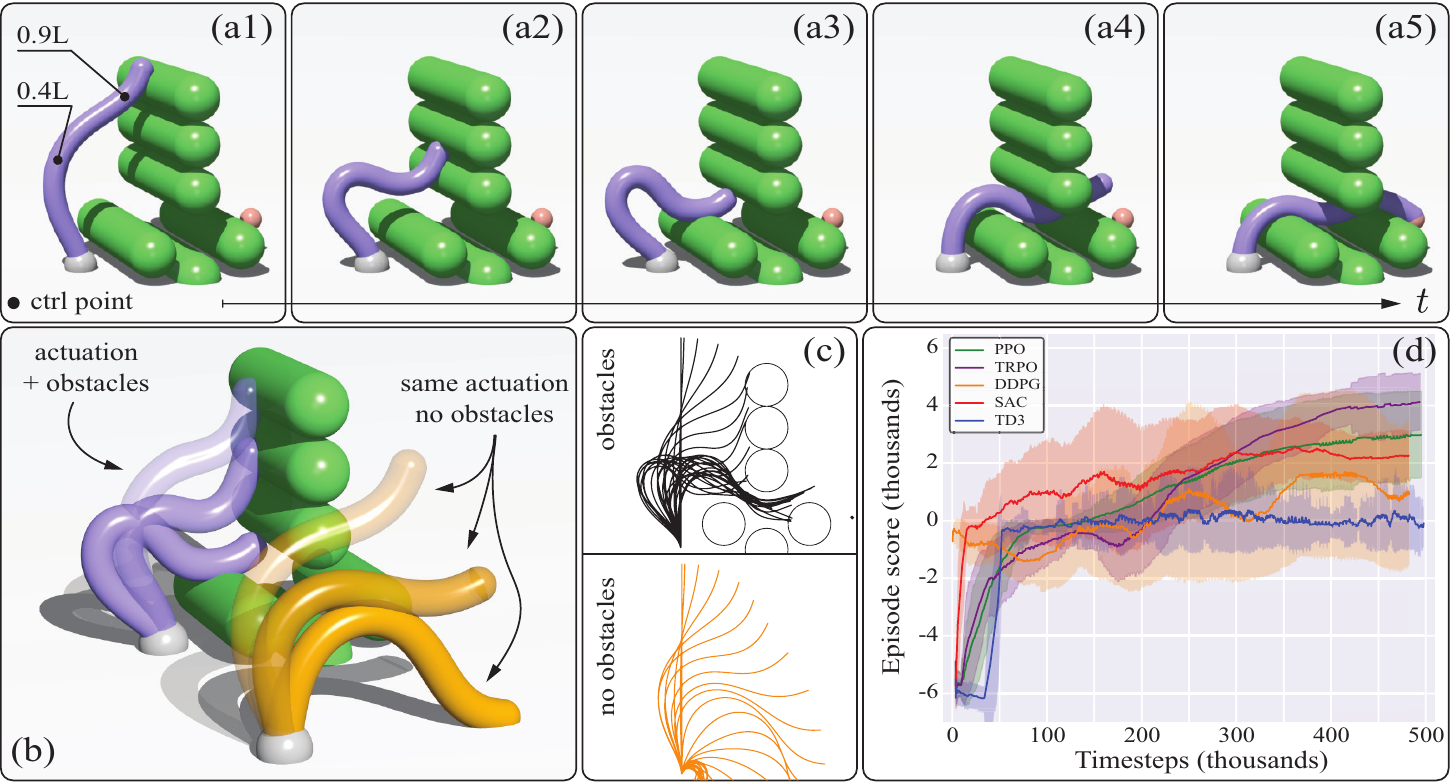}
\caption{a) Snapshots from a trained policy (TRPO) showing the arm leveraging the obstacles to maneuver through the opening and reach the target. b) Comparison of the behavior for the same applied actuation in the presence and absence of obstacles. c) Timelapse of centerlines of arm with and without obstacles showing how the interaction with obstacles is key to successfully maneuvering through the opening. d) Learning results of the different algorithms. Algorithms were trained with 5 different random seeds for 500 thousand timesteps. Curves are the rolling 50 sample average of the combined results and shaded regions are the standard deviation of the sample.}
\label{fig:prob3}
\vspace{-10pt}
\end{figure}

\textbf{Case 3 -- Underactutated maneuvering between structured obstacles.} A major advantage of compliant robots is their ability to maneuver around obstacles without damaging themselves or the obstacles. To explore the ability of model-free methods to learn to interact with and take advantage of solid boundaries, a stationary target is placed behind an array of obstacles with an opening through which the arm must reach (Figure \ref{fig:prob3}). The target is placed in the normal plane, so that only in-plane actuation is required. Obstacles and target locations are the same each episode and the reward is the same as Case 1. Importantly, no penalty is included to avoid contact with obstacles. Indeed, we do not see them as additional constraints, bur rather environmental features to be exploited. 

The obstacles are arranged so that it is not possible for the arm to fit through the opening without bending around or conforming to them. This results in a problem that cannot be solved by a rigid-link arm with a small number of DOFs but should be solvable by a compliant arm. To explore the interplay of underactuation and boundaries, only 2 control points manually placed at locations 0.4L and 0.9L along the arm are used. The rationale being that actuation at the mid-control point (0.4L) can organize an approximate global deformation sufficient to point the tip towards the opening, and subsequently push the arm in that general direction. Actuation at 0.9L helps navigating the obstacle morphology by bending the tip so as to determine along which surfaces the arm slides when pushed. 

The state information is the same as in Case 1 with the addition of the obstacle locations. Limited hyperparameter tuning was performed for off-policy methods while the hyperparameters from Case 2 were used for the on-policy methods. Because the target location is constant for each episode, only 500 thousand timesteps were necessary to train the policies. On-policy methods, TRPO in particular, were successful, as can be seen in Fig.~\ref{fig:prob3}a, extensively making use of boundaries to correct and redirect the imprecise actuation intrinsic in the use of only 2 DOFs, clearly a challenging and extremely underactuated setup. Off-policy methods were found to explore the action space too vigorously, slamming the arm against the obstacles, which led to numerical instabilities in the simulation that prevented them from always successfully learning to complete the task. 


\begin{figure}[t!]
\centering
\includegraphics[width=\textwidth]{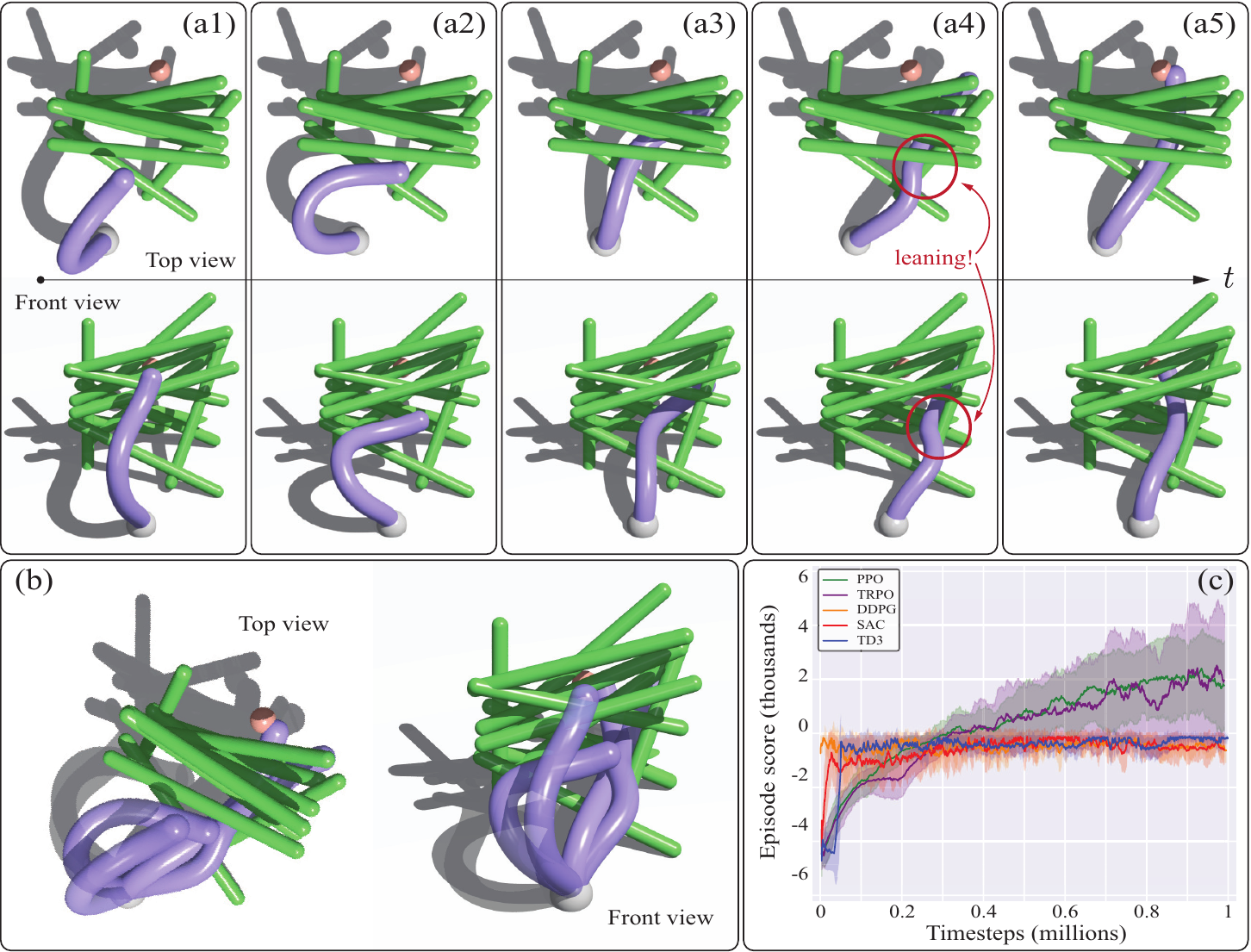}
\caption{a) Snapshots from a trained policy (TRPO) showing the arm successfully maneuvering through the obstacle nest to reach the target. b) Timelapse of arm maneuvering through obstacles. c) Learning results of the different algorithms. Algorithms were trained with 5 different random seeds for 1 million timesteps. Curves are the rolling 50 sample average of the combined results and shaded regions are the standard deviation of the sample.}
\label{fig:prob4}
\vspace{-10pt}
\end{figure}

\textbf{Case 4 -- Underactutated maneuvering between unstructured obstacles.} The final case expands on the arm's ability to interact with its environment by tasking it with finding its way through a nest of unstructured obstacles (Fig.~\ref{fig:prob4}a-b). The reward function, state, and actuation control points are the same as in Case 3. However, to navigate though the nest 3D bending is necessary, therefore internal torques are allowed to act in the normal and binormal directions, resulting in an action space with 4 DOFs. As with Case 3, minimal hyperparameter tuning was performed. Policies were trained for 1 million timesteps (Fig.~\ref{fig:prob4}c). Algorithm performance is similar to Case 3 with TRPO and PPO successfully learning to complete the task, while the off-policy methods are generally unable to select actions that allow them to remain numerically stable. As with Case 3, this problem is extremely challenging, if not impossible, for a rigid-link robot with such a few DOFs to solve, whereas our compliant arm is found to rather easily sneak though the nest by extensively leaning against various surfaces to redirect the tip towards the target.  

The key aspect of the underactuated control demonstrated here is the coupling of the compliant arm with its environment. A compliant robot can solve this problem with only two control points because of its ability to lean against and conform to obstacles. This is illustrated in Fig.~\ref{fig:prob3}b-c, which shows how the same action produces different arm behaviors when allowed to interact with the obstacles versus when obstacles are not considered. Further, Fig.~\ref{fig:prob4}a illustrates how the compliant arm leans against obstacles in order to maneuver itself through the obstacles. We note that when a traditional approach of avoiding obstacles via penalty terms is employed \cite{khatib1986real}, our flexible arm is unable to complete these tasks (see SI) as two DOFs do not provide the necessary finesse for the arm to maneuver without entering into contact with the solid boundaries. In contrast, because here the arm is allowed to interact with its environment, RL finds it natural to make use of boundaries as a resource, thus effectively simplifying the control problem. However, this is only possible when elastic effects are properly considered, demonstrating how the use of Elastica can spur the development of efficient control approaches for soft robots that make full use of their compliance, unlike traditional rigid-body physics simulators. 



\section{Conclusion}
\label{sec:conclusion}
	
	To fully realize the promised benefits of soft robots, it is necessary to develop control methods that exploit their unique physical properties. This is complicated by the difficulty of accurately modeling compliant structures in a simulation environment. Currently available simulation testbeds are insufficient for taking full advantage of elasticity in control. To address this, in this paper we introduced Elastica, a free, open-source physics environment for simulating assemblies of soft, slender, and compliant rods (as well as traditional rigid-body structures). Elastica can be interfaced with preexisting RL implementations (such as Stable Baselines) to straight-forwardly enable simulation-based learning for control of soft robots. We showed that a number of popular state-of-the-art RL methods (TRPO, PPO, DDPG, TD3, and SAC) are able to successfully learn to control a soft arm and to complete successively challenging tasks. We further demonstrate how the modeling of the arm's compliant mechanics, and its interaction with the environmental can help to simplify the control problem. The source code for Elastica and for the scenarios considered here is freely available online at \url{https://cosseratrods.org}, allowing these cases to serve as benchmarks for new algorithms that seek to control soft structures. Future work will seek to train dynamic control methods with Elastica that can be applied to real, physical robots.



\subsection*{Software Availability}
A open-source Python implementation of Elastica (PyElastica) is available online at \url{https://github.com/GazzolaLab/PyElastica} with documentation and examples available at \url{https:cosseratrods.org}.

\subsection*{Software Availability}
Financial and computational support for this work was provided by ONR MURI N00014-19-1-2373 (M.G., G.C.), NSF EFRI C3 SoRo \#1830881 (M.G.), NSF  CAREER  \#1846752  (M.G.), NSF/USDA \#2019-67021-28989 (M.G., G.C.), the Blue  Waters  project (OCI-0725070,  ACI-1238993) and XSEDE Stampede2 allocation TG-MCB190004 at the Texas Advanced Computing Center (TACC).

\clearpage



\bibliographystyle{ieeetr}
\bibliography{Elastica2020}  


\end{document}